%% file: decisiondb_manuscript.tex
\documentclass[11pt]{article}

\usepackage{amsmath}
\usepackage{amssymb}
\usepackage{graphicx}
\usepackage[margin=1in]{geometry}
\usepackage{tikz}
\usetikzlibrary{arrows.meta, positioning, calc, decorations.pathreplacing}
\usepackage{array}
\usepackage{booktabs}
\usepackage[table]{xcolor}

\newcolumntype{R}[1]{>{\raggedright\arraybackslash}p{#1}}
\usepackage[font=small]{caption}
\usepackage{hyperref}
\usepackage[numbers]{natbib}

\definecolor{tableShade}{RGB}{246, 243, 251}
\newcommand{\tablestyle}{\rmfamily\small\renewcommand{\arraystretch}{1.35}}

\usepackage{sectsty}
\allsectionsfont{\rmfamily}

\title{On Decision-Valued Maps and Representational Dependence}
\author{Gil Raitses}
\date{}

\begin{document}

\maketitle

\input{sections/abstract}

\input{sections/introduction}

\input{sections/problem_scope}

\input{sections/core_object}

\input{sections/protocol}

\input{sections/system_design}

\input{sections/empirical}

\input{sections/related_work}

\input{sections/limitations}

\input{sections/conclusion}

\small
\setlength{\bibsep}{2pt plus 1pt minus 1pt}
\bibliographystyle{plainnat}
\bibliography{bibliography}

\end{document}

%% file: sections/abstract.tex
\begin{abstract}
A computational engine applied to different representations of the same data can produce different discrete outcomes, with some representations preserving the result and others changing it entirely. A decision-valued map records which representations preserve the outcome and which change it, associating each member of a declared representation family with the discrete result it produces. This paper formalizes decision-valued maps and describes DecisionDB, an infrastructure that logs, replays and audits these relationships using identifiers computed from content and artifacts stored in write-once form. Deterministic replay recovers each recorded decision identifier exactly from stored artifacts, with all three identifying fields matching their persisted values. The contribution partitions representation space into persistence regions and boundaries, and treats decision reuse as a mechanically checkable condition.
\end{abstract}

%% file: sections/introduction.tex
\section{Introduction}

Analytical pipelines produce discrete outcomes that depend on how their inputs are represented. The same data, processed through the same computational engine under the same query, can yield different outcomes when the internal representation changes. Some representational changes leave the outcome intact; others alter it entirely.

In current practice, a pipeline runs under one representation, a result is reported, the sensitivity of that result to representational alternatives remains invisible. Pipeline outputs provide no basis for distinguishing decisions that can be reused under representational change from those requiring renewed justification.

A \emph{decision-valued map} records which representations preserve the outcome and which change it, associating each member of a representation family with the discrete result the engine produces. Figure~\ref{fig:pipeline} summarizes the evaluation protocol used to construct such a map by varying representations within a single arena.

Every stage of the evaluation chain is logged as a content-addressed artifact stored in write-once form by DecisionDB, the system described in this paper. DecisionDB supports representational sweeps through systematic variation of declared representation parameters, replay verification through deterministic recovery of decision identifiers from persisted artifacts and post-hoc audit of the full provenance chain.

In a graph routing demonstration, two representation parameters that control edge-cost construction are swept while the graph snapshot and shortest-path engine remain constant. One parameter preserves decision identity across its tested range; the other induces a discrete identity change at a specific threshold. Replay verification recovers all persisted identifiers exactly.

%% file: sections/problem_scope.tex
\section{Problem Scope}

Decision-valued mapping addresses systems with the following structure. A \emph{snapshot} $s$ is a frozen slice of external inputs over a declared time window; any change to the world state produces a new snapshot. A \emph{representation} $r \in \mathcal{R}(s)$ is a deterministic encoding of $s$, defined by explicit structural choices such as kernels, thresholds, weighting rules and aggregation policies; each representation is fully specified by a declared parameter set and generated by a versioned factory. An \emph{engine} $E$ is a computational procedure that consumes a representation and produces raw output, with engine configuration and version held constant during analysis. An \emph{equivalence policy} $\pi$ is a declared rule that reduces raw engine output to a discrete \emph{decision identity} $d \in \mathcal{D}$, defining when two raw outputs correspond to the same identity independent of incidental numerical differences.

Varying the representation while holding the snapshot, engine and query constant reveals where decision identity persists and where it changes. The central question is whether a previously produced decision remains admissible when the representation that supported it is altered. Persistence and boundary formation are treated as properties of the computational setting. Continuous outputs are in scope only when reduced to discrete identities via a declared policy. The map does not introduce training procedures, adaptive updates, gradient-based optimization or online learning.

This framing applies wherever discrete outcomes emerge from complex pipelines and representational choices may influence those outcomes. Routing evaluated under alternative cost encodings, classification evaluated under alternative feature constructions and resource allocation evaluated under alternative aggregation rules all exhibit the same structural question.

%% file: sections/core_object.tex
\section{The Decision-Valued Map}

The central object of study is a mapping
\[
f \colon \mathcal{R} \to \mathcal{D},
\]
where $\mathcal{R}$ denotes a family of representations over a fixed snapshot $s$ and $\mathcal{D}$ denotes a set of discrete decision identities. For each representation $r \in \mathcal{R}$, the engine $E$ produces raw output $E(r)$, and the equivalence policy $\pi$ extracts a decision identity $d = \pi(E(r))$.

Three structural features of this map are observable through controlled variation of $\mathcal{R}$. Persistence regions are connected subsets of $\mathcal{R}$ over which $f$ is constant; within a persistence region, representational variation preserves the outcome. Boundaries are loci in $\mathcal{R}$ where $f$ changes value, separating two persistence regions with different decision identities. Fractures are boundaries where a small change in representation parameters induces a discrete identity change, indicating high sensitivity of the outcome to the representation.

The purpose of DecisionDB is to \emph{materialize} $f$, evaluating it at declared points in $\mathcal{R}$, storing the results as immutable artifacts, then making the resulting map queryable, replayable and auditable. Figure~\ref{fig:pipeline} illustrates this protocol.

\definecolor{erdHeader}{HTML}{e1d5e7}
\definecolor{erdStroke}{HTML}{b5739d}
\definecolor{arenaFill}{HTML}{f3eff7}

\begin{figure}[t]
  \centering
  \begin{tikzpicture}[
    box/.style={draw=erdStroke, rounded corners=4pt, align=center, inner sep=5pt, minimum height=13mm, font=\rmfamily\footnotesize, fill=erdHeader},
    smallbox/.style={draw=erdStroke, rounded corners=3pt, align=center, inner sep=4pt, minimum height=10mm, font=\rmfamily\scriptsize, fill=erdHeader!40},
    idbox/.style={draw=erdStroke, rounded corners=3pt, align=center, inner sep=3pt, minimum height=8mm, font=\rmfamily\scriptsize},
    arrow/.style={-{Latex[length=2.5mm, width=2mm]}, thick, erdStroke},
    eqlbl/.style={font=\rmfamily\scriptsize, erdStroke},
    arenalbl/.style={font=\rmfamily\scriptsize, text=black}
  ]

  \draw[erdStroke!40, rounded corners=6pt, fill=arenaFill] (-0.8, 1.0) rectangle (4.6, -0.6);
  \node[arenalbl, anchor=south west] at (-0.8, 1.05) {arena (invariant)};

  \node[box, text width=20mm] (snap) at (0.5, 0.2) {Snapshot $s$};
  \node[box, text width=20mm] (eng) at (3.3, 0.2) {Engine $E$};

  \node[smallbox, text width=22mm] (r1) at (7.0, 1.6) {$r_1$};
  \node[smallbox, text width=22mm] (r2) at (7.0, 0.2) {$r_2$};
  \node[smallbox, text width=22mm] (r3) at (7.0, -1.2) {$r_3$};
  \node[arenalbl, anchor=south] at (7.0, 2.2) {representation family $\mathcal{R}$};

  \draw[arrow] (eng.east) -- ++(0.4,0) |- (r1.west);
  \draw[arrow] (eng.east) -- (r2.west);
  \draw[arrow] (eng.east) -- ++(0.4,0) |- (r3.west);

  \draw[blue!8, rounded corners=4pt, fill=blue!6] (9.5, 2.1) rectangle (11.0, -0.3);
  \node[idbox, fill=blue!12] (d1) at (10.2, 1.6) {$d_A$};
  \node[idbox, fill=blue!12] (d2) at (10.2, 0.2) {$d_A$};
  \node[idbox, fill=orange!18] (d3) at (10.2, -1.2) {$d_B$};

  \draw[arrow] (r1.east) -- (d1.west);
  \draw[arrow] (r2.east) -- (d2.west);
  \draw[arrow] (r3.east) -- (d3.west);

  \node[arenalbl, anchor=west] at (11.2, 0.9) {persistence};
  \node[arenalbl, anchor=west] at (11.2, -1.2) {boundary};

  \end{tikzpicture}
  \caption{Evaluation protocol for a decision-valued map. A single arena, consisting of a data snapshot and a computational engine, is reused across evaluation. Representations drawn from a declared family are varied within this arena and evaluated independently. Agreement of decision identifiers across representations indicates persistence of decision identity; a change in identifier indicates a boundary induced by representational variation alone.}
  \label{fig:pipeline}
\end{figure}
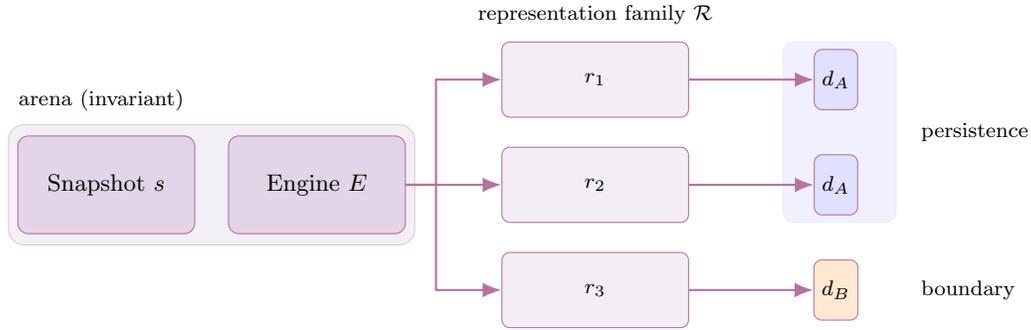

%% file: sections/protocol.tex
\section{Sweep Protocol}

A \emph{representational sweep} evaluates the decision-valued map $f$ across a declared set of representations. The protocol proceeds through five stages, summarized in Table~\ref{tab:sweep-protocol}, that transform a frozen snapshot into a queryable decision map.

A sweep begins by freezing a snapshot and assigning it a content-addressed identifier. A representation family is then declared via a deterministic factory, and a sweep plan specifies the parameter grid, engine and equivalence policy. The engine executes independently for each representation, producing raw outputs stored as immutable artifacts. The equivalence policy reduces each output to a decision identity, and the decision map table records the links from snapshot and representation through engine run to decision identity. All artifacts are versioned and linked through content-addressed identifiers, so the resulting materialized map supports reproducible replay and post-hoc analysis without re-executing the engine.

\begin{table}[t]
\centering
\caption{Five-stage representational sweep protocol.}
\label{tab:sweep-protocol}
\smallskip
\tablestyle
\rowcolors{2}{tableShade}{white}
\begin{tabular}{@{\hskip 6pt}c l R{40mm} R{42mm}@{\hskip 6pt}}
\toprule
\rowcolor{white}
 & Stage & Inputs & Outputs \\
\midrule
1 & Freeze snapshot & World state, time window & Snapshot identifier \\
2 & Declare representations & Snapshot ref, parameterization, factory version & Representation identifiers, one per parameter setting \\
3 & Plan sweep & Parameter grid, engine name and version, equivalence policy & Sweep plan descriptor, itself content-addressed \\
4 & Execute engine & Each representation independently & Run records, raw output artifacts stored as immutable files \\
5 & Extract decisions & Raw outputs, equivalence policy $\pi$ & Decision identifiers, decision map entries linking representations and runs to decisions \\
\bottomrule
\end{tabular}
\end{table}

%% file: sections/system_design.tex
\section{System Design}

DecisionDB is implemented as a Python package backed by SQLite. It manages five entity types through a relational schema with content-addressed primary keys and foreign-key constraints.

\subsection{Content Addressing}

All identifiers are computed deterministically from content. Given an entity's payload as a Python dictionary, DecisionDB serializes it to canonical JSON with keys sorted alphabetically, no whitespace, arrays in declaration order, floats as strings and explicit version fields. It then computes the SHA-256 digest of the UTF-8 encoding, truncates to the first 16 hexadecimal characters and prepends a type-specific prefix. Identical content always produces identical identifiers, regardless of when or where the computation occurs.

\subsection{Schema}

\begin{figure}[t]
  \centering
  \includegraphics[width=0.95\textwidth]{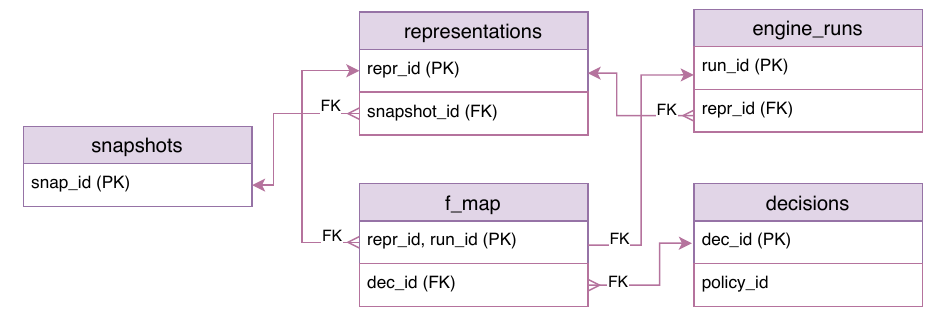}
  \caption{DecisionDB relational schema. Five tables form a content-addressed provenance chain. Foreign-key arrows indicate the direction of referential dependency, linking representations to their parent snapshot, engine runs to the representation consumed, the decision map table to representations, runs and decision identities.}
  \label{fig:schema}
\end{figure}

The relational schema contains five core tables. Figure~\ref{fig:schema} shows their foreign-key relationships and Table~\ref{tab:schema} summarizes each table's role. Each table enforces content-addressed primary keys and foreign-key constraints that link the full provenance chain from snapshot through representation and engine execution to decision identity. All writes are append-only, scoped by experiment identifier and executed within transactions. Inserts use an insert-or-ignore strategy for idempotency, so re-inserting the same content-addressed entity is a no-op.

\begin{table}[t]
\centering
\caption{DecisionDB entity roles.}
\label{tab:schema}
\smallskip
\tablestyle
\rowcolors{2}{tableShade}{white}
\begin{tabular}{@{\hskip 8pt}l l R{68mm}@{\hskip 8pt}}
\toprule
\rowcolor{white}
Table & Key prefix & Role \\
\midrule
snapshots & snap & Immutable input state and its artifact manifest \\
representations & repr & Deterministic encodings of snapshots under a declared parameterization \\
engine runs & run & Execution records of the fixed engine on specific representations \\
decisions & dec & Discrete decision identities extracted by an equivalence policy \\
f map & composite & Materialized decision-valued map linking representations, runs and decisions \\
\bottomrule
\end{tabular}
\end{table}

\subsection{Equivalence Policies}

An equivalence policy defines how raw engine output is reduced to a decision identity. Each policy specifies a hash source, identifying which field of the raw output carries decision-relevant content such as the route node sequence. It also specifies a canonicalization rule that determines how to serialize the extracted content alongside a match rule that determines identity such as SHA-256 equality. The policy itself is content-addressed, so any change to the policy definition produces a new policy identifier and new decision identifiers downstream.

\subsection{Replay Verification}

Replay verification takes a persisted decision record, reloads the stored raw output and policy specification, recomputes the policy identifier, payload hash, decision identifier, then checks them against the persisted values. Because replay is read-only and writes no rows, a successful pass confirms that the content-addressing chain from raw output through policy application to decision identity is deterministic and self-consistent.

%% file: sections/empirical.tex
\section{Empirical Demonstration}

A graph routing problem provides the concrete setting for demonstrating how the decision-valued map makes representational dependence observable.

\subsection{Setup}

The demonstration fixes a directed graph with node set $V$ where $|V| = 564$, edge set $E$ and immutable edge attributes including baseline costs derived from geographic distance and a normalized stress metric. The snapshot is content-addressed and sealed before any engine execution. A single origin-destination query is fixed at start node 85 and end node 50.

Each representation encodes the graph's edge costs as a deterministic function of the fixed edge attributes. Two representation parameters control the cost surface. The first, neighbor weight, applies a weight to a neighbor-based cost component and is tested at 0.5 and 1.0. The second, second-order weight, applies a weight to a second-order cost component and is tested at 0.25 and 0.5. Each sweep varies one parameter while holding the other fixed, producing two representation variants per sweep and four engine evaluations total.

The engine is a shortest-path solver using Dijkstra's algorithm~\cite{dijkstra1959note}, with configuration and version held constant across all runs. Execution times ranged from 0.5 to 1.4 milliseconds.

Two routes are assigned the same decision identity if and only if they traverse identical node sequences. The equivalence policy enforces this by computing an identifier from a JSON-serialized, sorted-key and UTF-8-encoded representation of the route nodes.

\subsection{Results}

In this instantiation, decision identities correspond to route equivalence classes, but the protocol applies to any discrete outcome produced by a fixed decision query.

The neighbor weight sweep varies the neighbor weight from 0.5 to 1.0 while holding second-order weight fixed at 0.25. Both representations yield the same 16-node path, Decision~A, indicating persistence across the tested range.

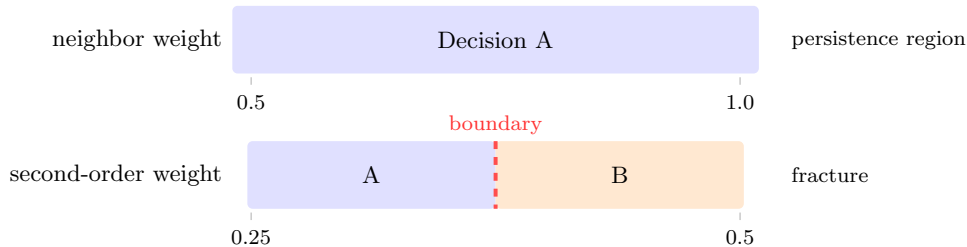
\begin{figure}[t]
  \centering
  \begin{tikzpicture}[
    region/.style={minimum height=9mm, align=center, font=\rmfamily\footnotesize, rounded corners=2pt},
    regionA/.style={region, fill=blue!12},
    regionB/.style={region, fill=orange!18},
    lbl/.style={font=\rmfamily\footnotesize, anchor=east},
    tick/.style={font=\rmfamily\scriptsize, text=black},
    plbl/.style={font=\rmfamily\scriptsize, text=black, anchor=west}
  ]

  \node[lbl] at (-0.3, 0) {neighbor weight};
  \node[regionA, minimum width=70mm] at (3.2, 0) {Decision A};
  \node[tick, anchor=north] at (-0.05, -0.6) {0.5};
  \node[tick, anchor=north] at (6.45, -0.6) {1.0};
  \draw[black!30] (-0.05, -0.5) -- (-0.05, -0.6);
  \draw[black!30] (6.45, -0.5) -- (6.45, -0.6);
  \node[plbl] at (7.0, 0) {persistence region};

  \node[lbl] at (-0.3, -1.8) {second-order weight};
  \node[regionA, minimum width=33mm] at (1.55, -1.8) {A};
  \node[regionB, minimum width=33mm] at (4.85, -1.8) {B};
  \node[tick, anchor=north] at (-0.05, -2.4) {0.25};
  \node[tick, anchor=north] at (6.45, -2.4) {0.5};
  \draw[black!30] (-0.05, -2.3) -- (-0.05, -2.4);
  \draw[black!30] (6.45, -2.3) -- (6.45, -2.4);
  \node[plbl] at (7.0, -1.8) {fracture};

  \draw[red!70, line width=1.5pt, dashed] (3.2, -1.35) -- (3.2, -2.25);
  \node[font=\rmfamily\scriptsize, text=red!80, fill=white, inner sep=1pt, anchor=south] at (3.2, -1.3) {boundary};

  \end{tikzpicture}
  \caption{Identity persistence and fracture structure for the two representation parameters. The neighbor weight parameter spans a single persistence region in which both tested values produce Decision~A, indicating decision identity is stable across this range. The second-order weight parameter spans two regions separated by a fracture, where decision identity changes from Decision~A to Decision~B between the values 0.25 and 0.5. The exact threshold is not resolved by this two-point sample.}
  \label{fig:persistence-fracture}
\end{figure}

The second-order weight sweep varies the second-order weight from 0.25 to 0.5 while holding neighbor weight fixed at 0.5. At second-order weight 0.25, the engine produces Decision~A. At second-order weight 0.5, the engine produces Decision~B, a distinct 14-node path traversing a different region of the graph. The boundary between these two decision identities lies between second-order weight values 0.25 and 0.5, where a small parameter change induces a qualitative change in the discrete outcome. Full node sequences for each decision are stored as raw output artifacts and are recoverable via replay verification. Table~\ref{tab:sweep-summary} summarizes the sweep results; Figure~\ref{fig:persistence-fracture} shows the persistence and fracture structure.

\begin{table}[t]
\centering
\caption{Representational sweep results.}
\label{tab:sweep-summary}
\smallskip
\tablestyle
\rowcolors{2}{tableShade}{white}
\begin{tabular}{@{\hskip 8pt}l l c c l@{\hskip 8pt}}
\toprule
\rowcolor{white}
Sweep parameter & Value & Decision & Nodes & Boundary \\
\midrule
neighbor weight & 0.5 & A & 16 & \\
neighbor weight & 1.0 & A & 16 & No \\
second-order weight & 0.25 & A & 16 & \\
second-order weight & 0.5 & B & 14 & Yes \\
\bottomrule
\end{tabular}
\end{table}

\subsection{Replay Verification}

Replay verification is performed on a decision produced by the sweep. The procedure reloads the stored raw output and policy specification, recomputes all three identifying fields so as to check them against the persisted values. All recomputed values match exactly, as shown in Table~\ref{tab:replay}.

\begin{table}[t]
\centering
\caption{Replay verification results.}
\label{tab:replay}
\smallskip
\tablestyle
\rowcolors{2}{tableShade}{white}
\begin{tabular}{@{\hskip 8pt}l l l@{\hskip 8pt}}
\toprule
\rowcolor{white}
Field & Persisted & Recomputed \\
\midrule
Policy ID & pol\_d8da3e00e9584eb1 & pol\_d8da3e00e9584eb1 \\
Payload hash & 3a9d63ac28378116 & 3a9d63ac28378116 \\
Decision ID & dec\_e28092c4dc33b8f1 & dec\_e28092c4dc33b8f1 \\
\bottomrule
\end{tabular}
\end{table}

The replay is read-only and leaves the database state unchanged. Table counts before and after replay are identical, with 1~engine run, 1~decision, 1~f\_map entry. This confirms that the content-addressing chain from raw output through policy application to decision identity is deterministic and end-to-end auditable.

%% file: sections/related_work.tex
\section{Related Work}

The sensitivity of analytical conclusions to representational and analytic choices has been documented across multiple disciplines. This section situates the decision-valued mapping framework relative to existing work on analytic flexibility, reproducibility infrastructure and provenance systems. Table~\ref{tab:comparison} summarizes the key distinctions.

\subsection{Analytic Flexibility and Multiverse Methods}

The dependence of statistical conclusions on analytic choices has been examined through several complementary lenses. Specification curve analysis~\cite{simonsohn2020specification} evaluates how reported effects vary across defensible analytic specifications. Multiverse analysis~\cite{steegen2016increasing} extends this by jointly varying data processing and model specification decisions to characterize the full space of results consistent with a dataset. The same instability appears even when researchers follow pre-registered protocols, through implicit degrees of freedom that shape findings absent deliberate p-hacking~\cite{gelman2013garden}. The scale of this effect in observational studies is substantial, with effect estimates fluctuating across model specifications~\cite{patel2015assessment}.

Decision-valued mapping shares the premise that analytic conclusions depend on choices that are often left implicit. Specification curve and multiverse analyses operate on continuous effect estimates such as regression coefficients and p-values, then visualize their distribution. Decision-valued maps operate on discrete outcome identities and characterize the topology of representation space, identifying which regions preserve identity and where boundaries form. The result is a partition of representation space, in contrast to the distributions produced by specification curve and multiverse analyses.

\subsection{Sensitivity Analysis}

Sensitivity analysis quantifies how variation in model inputs contributes to variation in model outputs, typically through variance decomposition or derivative-based indices over continuous output spaces~\cite{saltelli2008global}. Where sensitivity analysis decomposes output variance and computes sensitivity indices, decision-valued mapping directly observes whether a discrete outcome changes or persists under representation variation. Decision-valued mapping is compatible with sensitivity analysis, since sensitivity indices could be computed over a binarized decision map, but it does not require or assume a continuous output metric.

\subsection{Reproducibility in Machine Learning}

Pipelines satisfying identical training criteria can yield predictors with divergent behavior under distribution shift~\cite{damour2022underspecification}. Deep reinforcement learning results are sensitive to implementation details and hyperparameter choices in ways that standard reporting obscures~\cite{henderson2018deep}, and variance accounting across machine learning benchmarks has been formalized to quantify this effect~\cite{bouthillier2021accounting}. Reproducibility checklists~\cite{pineau2021improving} encourage reporting of experimental details but omit content-addressed provenance while omitting deterministic replay.

Decision-valued mapping addresses a related but distinct problem. Where reproducibility work documents variability across training runs and hyperparameter settings, decision-valued mapping holds snapshot and engine invariant while varying representation and tracking its effect on discrete outcome identity. Decision-valued mapping complements reproducibility checklists by providing a lower-level infrastructure for testing whether specific outcomes are stable under specific representational changes.

\subsection{Provenance and Workflow Systems}

{\sloppy
General-purpose provenance vocabularies record the derivation history of computational
artifacts~\cite{moreau2013prov}. Workflow provenance for scientific computations~\cite{freire2012provenance}, experiment tracking for machine learning pipelines~\cite{zaharia2018accelerating}, standardized workflow definitions for reproducible execution~\cite{amstutz2016common} and lineage tracking for fault-tolerant distributed computation~\cite{zaharia2012resilient} all address aspects of this problem.
\par}

The diagnostic layer described here is narrower than these systems in scope and more specific in its invariants. Where general-purpose provenance systems manage arbitrary workflows and track general-purpose provenance graphs, this layer enforces a specific structure with immutable snapshots, declared representation families, fixed engines and equivalence policies that reduce raw output to discrete identities. The content-addressing scheme ensures that identical inputs always produce identical identifiers, and replay verification checks end-to-end consistency of the provenance chain. The specificity of this structure yields a queryable diagnostic object in the decision-valued map, a capability that general-purpose provenance systems lack direct support for.

\begin{table}[t]
\centering
\caption{Summary comparison between decision-valued mapping and related approaches.}
\label{tab:comparison}
\smallskip
\tablestyle
\rowcolors{2}{tableShade}{white}
\begin{tabular}{@{\hskip 6pt}R{30mm} R{40mm} R{44mm}@{\hskip 6pt}}
\toprule
\rowcolor{white}
Dimension & Typical approaches & Decision-valued mapping \\
\midrule
Object of analysis & Continuous effect estimates, variance, p-values & Discrete decision identities \\
Variation source & Model specs, preprocessing, hyperparameters & Representation parameters under fixed snapshot and engine \\
Output characterization & Distributions, sensitivity indices, specification curves & Persistence regions, boundaries, fractures \\
Infrastructure & Logging, provenance metadata, workflow graphs & Content-addressed, replayable decision maps \\
\bottomrule
\end{tabular}
\end{table}

\subsection{Infrastructural Precedents}

The approach follows a pattern observed in the development of foundational abstractions. Abstract data types~\cite{liskov1974programming} separated representation from observable behavior, enabling systems to evolve internally while preserving external guarantees. Write-ahead logging~\cite{gray1993transaction} transformed durability from an ad-hoc property into an auditable, replayable state transition protocol. In both cases, a diagnostic layer that made structural dependence explicit and testable constituted the main contribution.

Each decision occurs within an arena defined by a snapshot, an engine and a query. The protocol records how outcome identity evolves across representational variants within that arena. Methods that adapt model parameters during evaluation concentrate effort on producing stronger solutions within that arena. The present protocol instead addresses reuse, asking whether the recorded context is sufficient to justify carrying a discrete outcome across representations and downstream consumers. Decision-valued mapping extends this lineage to analytical pipelines whose outputs are discrete, consequential and contingent on representational choice.

%% file: sections/limitations.tex
\section{Limitations}

Decision-valued mapping applies to systems whose outputs can be reduced to discrete identities via a declared equivalence policy. Continuous outputs such as probability distributions and regression surfaces are out of scope unless such a reduction is explicitly defined. The choice of equivalence policy directly determines what counts as ``the same outcome,'' and different policies applied to the same raw output will in general produce different decision maps.

DecisionDB does not track how decision maps evolve across snapshot or engine versions. Results reported here assume a single snapshot and engine held invariant, so each snapshot-engine combination requires a separate analysis. Any change to the input data, model parameters, or execution logic constitutes a new analytical context.

The reported sweeps cover two representation parameters, neighbor weight at values 0.5 and 1.0, and second-order weight at values 0.25 and 0.5, applied to a single graph snapshot with a single origin-destination pair. The representation space is sampled at four points total, and unobserved regions remain unconstrained.

Persistence regions and boundaries identified here may not generalize to finer parameter grids, different origin-destination pairs, or different graph topologies.

The empirical demonstration uses graph routing. The approach is designed to be domain-agnostic, but it has not been validated on classification, resource allocation, or other pipeline types. Applying decision-valued mapping to a new domain requires defining an appropriate equivalence policy, which involves domain-specific judgment about what constitutes ``the same outcome.''

Observing that a boundary exists between two parameter values does not explain why it exists. The framework identifies where decision identity changes under representational variation but does not attribute the change to any specific mechanism within the engine or the representation construction.

The current implementation uses SQLite and has been tested with single-digit representation families. Scaling to large parameter grids of hundreds or thousands of representations would require evaluation of storage, query performance and sweep orchestration, none of which has been performed.

%% file: sections/conclusion.tex
\section{Conclusion}

In a graph routing demonstration, one representation parameter preserves decision identity across its tested range while another induces a discrete identity change. DecisionDB materializes this structure through a five-stage sweep protocol, a five-table relational schema and a replay verification procedure that recovers all persisted identifiers exactly.

The approach is limited by its restriction to discrete outcomes, its reliance on held-constant snapshots and engines, the narrow empirical coverage reported here. A natural extension is to densify the parameter sweep, apply the same protocol across additional domains, introduce a cross-snapshot comparison procedure that treats ``same query, different snapshot'' as an explicit axis of reuse admissibility.

The contribution of this work is infrastructural, introducing a diagnostic layer that records how discrete decisions arise from families of representations. Decision-valued maps expose stability and fracture directly, placing explicit representational context at the center of downstream reuse decisions, with confidence or performance summaries in a complementary role. Decision reuse becomes a compatibility question that can be checked mechanically before deployment across tasks and timescales.